%% file: neurips_2026.tex
\documentclass{article}

\usepackage[main,preprint]{neurips_2026}

\usepackage[utf8]{inputenc} 
\usepackage[T1]{fontenc}    
\usepackage{hyperref}       
\usepackage{url}            
\usepackage{booktabs}       
\usepackage{amsfonts}       
\usepackage{nicefrac}       
\usepackage{microtype}      
\usepackage{xcolor}         

\usepackage{amsmath}
\usepackage{multicol}
\usepackage{multirow}
\usepackage{tabularx}
\usepackage{xspace}
\usepackage{adjustbox}
\usepackage[table]{xcolor}
\usepackage{arydshln}
\usepackage{booktabs}
\usepackage{caption}
\newcolumntype{Y}{>{\Centering\arraybackslash}X}
\newcommand{\fullmodel}{\textbf{H}ybrid \textbf{T}oolset \textbf{A}gentization \& \textbf{A}daptation \xspace}
\newcommand{\model}{HTAA\xspace}
\newcommand{\dataname}{InfoVerify\xspace}

\newcommand{\ignore}[1]{}

\definecolor{lightergray}{RGB}{230,230,230}
\definecolor{DarkGreen}{RGB}{30,130,30}
\usepackage{wrapfig}
\usepackage{marvosym}
\usepackage{pifont}
\usepackage{xcolor}
\usepackage{algorithm}
\usepackage{algpseudocode}

\title{HTAA: Enhancing LLM Planning via Hybrid Toolset Agentization \& Adaptation }

\author{%
  Chengrui Huang$^{1}$\thanks{The first two authors contributed equally.~~~\Letter~Corresponding author.} Junshuo Zhang$^{1*}$  Zhiyuan Ma$^1$ Xikun Wang$^{2}$\\ 
  \textbf{Ximeng Wang$^{2}$} \textbf{Menghua Jiang$^{2}$} \textbf{Gang Zeng$^{2}$} \textbf{Zhaobing Han$^{2}$}\\ \textbf{Shen Gao$^{1}$}$\text{\Letter}$ \textbf{Shuo Shang$^{1}$}$\text{\Letter}$ \\
  $^{1}$ University of Electronic Science and Technology of China,\\
  $^2$DiDi Global Inc. \\
  \texttt{ry.cr.huang@gmail.com} \\
    \texttt{shengao@uestc.edu.cn},
    \texttt{jedi.shang@gmail.com}
}

\begin{document}

\maketitle

\begin{abstract}
Enabling large language models to scale and reliably use hundreds of tools is critical for real-world applications, yet challenging due to the inefficiency and error accumulation inherent in flat tool-calling architectures. 
To address this, we propose \fullmodel (\model), a hierarchical framework for scalable tool-use planning. 
We propose a novel toolset agentization paradigm, which encapsulates frequently co-used tools into specialized agent tools, thereby reducing the planner's action space and mitigating redundancy. 
To ensure effective coordination, we design Asymmetric Planner Adaptation, a trajectory-based training paradigm that aligns the high-level planner with agent tools via backward reconstruction and forward refinement.
To validate the performance of \model, we conduct experiments on a real-world internal dataset, \dataname, based on the POI validation workflow of China's largest online large-scale ride-hailing platform, featuring long-horizon executable tool trajectories. 
Experiments on \dataname and widely-used benchmarks show that \model consistently achieves higher task success rates, requires short tool calling trajectories, and significantly reduces context overhead compared to strong baselines. 
Furthermore, in a production deployment, \model substantially reduces manual validation effort and operational cost, demonstrating its practical efficacy.
\end{abstract}

\input{Sections/01-Introduction}

\input{Sections/02-Related-work}
\input{Sections/03-Method}
\input{Sections/04-Experiments}
\input{Sections/05-Conclusion}

\bibliographystyle{plainnat}
\bibliography{custom}
\input{Sections/06-Appendix}

\newpage
\input{checklist.tex}

\end{document}

%% file: Sections/01-Introduction.tex
\section{Introduction}

In recent years, large language models (LLMs) have demonstrated remarkable capabilities across a wide range of domains~\citep{wang2025cesrecconstructingpseudointeractions, feng-etal-2025-culfit}. 
Moreover, equipping LLMs~\citep{gpt4, touvron2023llama2openfoundation} with the ability to interact with external environments is crucial for enhancing their capacity to address complex real-world challenges, such as querying search engines to retrieve up-to-date information~\citep{NEURIPS2024_e4c61f57} or autonomously generating comprehensive travel itineraries~\citep{hao2024large,xie2024travelplanner}.
As LLMs advance, integrating external tools is crucial both for addressing real-world user requirements and for progressing toward artificial general intelligence~\citep{wang2023mint,liu2023agentbench,tian2024opportunities}

However, enabling LLMs to scalably and reliably use a large number of tools remains an open challenge. 
Most existing tool-augmented work~\citep{react,toolformer,patil2023gorilla} adopts a flat tool-calling architecture, where the planner directly selects from all available tools and explicitly schedules each invocation step.
These methods expose every tool to the planner and place the full burden of managing tool executions on it, leading to increasing complexity. 
As the toolset grows, the action space expands rapidly, and even moderately complex tasks may require long sequences of tool invocations. 
In this setting, the planner must simultaneously engage in task-level decision making and manage the order of each individual tool invocation, substantially increasing the planning burden and causing error accumulation. 
Thus, the first challenge is that \textbf{flat planner architecture usually suffers an error accumulation problem} when facing complex tasks and scaling to large toolsets.
Moreover, many tools provide overlapping or complementary functionalities~\citep{toolllm}. 
Invoking them independently forces the planner to process redundant outputs and repeatedly perform similar reasoning steps, which increases latency, lengthens context, and reduces overall efficiency. 
The second challenge is that \textbf{independently invoking each tool results in overall inefficiency}.
These challenges make it difficult to scale to real-world tool ecosystems containing hundreds of heterogeneous tools.

An intuitive idea is to employ a hierarchical approach, similar to the management structure of human organizations, where the highest-level decision-maker does not directly handle the detailed work but delegates it to lower-levels.
Introducing this strategy into complex tool-calling applications will allow the planner to operate within a smaller action space, focusing on task-level decisions. 
While this reduces the planner’s workload, it also hides the detailed outputs of tools, which may lead to discrepancies between the high-level planner and the low-level tool execution agent.

To address this, we propose \fullmodel (\model), a hierarchical framework designed to enhance the scalability and reliability of tool-augmented planning.
First, we introduce a tool abstraction mechanism called \textbf{toolset agentization}. 
It groups multiple frequently co-used or functionally overlapping tools under a single \textbf{agent tool}, allowing the higher-level planner to interact only with this agent tool.
Unlike existing approaches that expose every fine-grained function to the planner, each agent tool coordinates multiple tool calls and aggregates their results into a concise response. 
From the planner's perspective, completing a subtask requires only one interaction with these agent tools to obtain the results. 
This approach directly shortens the planning cycle, reduces the number of explicit decisions, and improves the overall efficiency of the system when handling complex tasks. 
Furthermore, we retain some simple and independent tools as basic tools for the planner to use directly. 
Ultimately, the planner accomplishes complex tasks by mixing the use of both agent tools and basic tools.

Although toolset agentization simplifies the planner's action space, it introduces a coordination consistency problem: the planner cannot directly observe the raw outputs but only aggregated responses and must interact with agent tools whose internal processes are opaque.
Therefore, we propose an asymmetric planner adaptation method, which is a trajectory-based training paradigm. This asymmetric strategy stabilizes training and enables the planner to align with agent tool behaviors without modifying the agent tools themselves. 
We construct high-quality supervision signals via backward trajectory reconstruction to efficiently obtain viable tool-use paths. 
Then, through forward refinement, we generate correct reasoning trajectories. 
Finally, we optimize the planner using supervised fine-tuning.

To validate the effectiveness of our proposed \model in real-world online internet applications, we conduct experiments on a proprietary internal dataset, \dataname.
This dataset is collected from the POI validation workflow of China's largest online ride-hailing platform. 
Originally, the entire POI validation process was performed by a fully manual team, utilizing tools such as map applications, web searches, online review analysis, and phone calls. 
We aim to substantially reduce this fully manual workflow with a tool-use framework, such as \model. 
On average, each POI validation task requires the use of 10 tools, combined with reasoning based on the tool return values to determine whether the POI information is valid.

Extensive experiments on \dataname and widely-used benchmarks demonstrate that \model achieves state-of-the-art performance. 
Toolset agentization reduces planning complexity and redundancy, while asymmetric adaptation improves coordination and robustness.
The experimental results show that \model consistently requires fewer tool calls compared to flat architectures and achieves higher task success rates across diverse tool-augmented reasoning benchmarks.
Furthermore, using the same \model framework with minor domain adaptation on China's largest online ride-hailing platform, it replaced the original manual POI validation team, reducing the workload of manual verification by 84.5\% and cutting manual annotation costs by 81.25\%.

\noindent Our contributions are summarized as follows: 

\noindent $\bullet$ We propose \fullmodel, a hierarchical framework that improves tool-use scalability and coordination by integrating tool abstraction and planner adaptation.

\noindent $\bullet$ We introduce \textbf{T}oolset \textbf{A}bstraction, which encapsulates basic tools into agent tools, reducing planning complexity, shortening trajectories, and mitigating redundancy and context overhead.

\noindent $\bullet$ We introduce \textbf{A}symmetric \textbf{P}lanner \textbf{A}daptation, a training paradigm that bridges the coordination gap between the planner and agent tools through efficient trajectory construction and stable optimization. 

\noindent $\bullet$ Experiments across multiple benchmarks demonstrate that \model consistently improves both effectiveness and efficiency, achieving higher task success rates while requiring shorter trajectories and incurring substantially lower context overhead than strong baselines.

%% file: Sections/02-Related-work.tex
\section{Related work}
\paragraph{Tool Learning}
Tool learning aims to expand the capability boundaries of LLMs by equipping them with external tools. Given a problem and a set of candidate tools, LLMs often need to interact with the environment multiple times to gather information by invoking a variety of tools. Many studies aim to achieve greater robustness from different perspectives. For instance, ToolLLM fine-tunes LLMs using expert trajectories synthesized via a depth-first search–based decision tree~\citep{toolllm}. Confucius~\citep{Confucius} proposes a curriculum learning approach to teach LLMs to master tools of varying difficulty levels. TP-LLaMA~\citep{tpllama} and StepTool~\citep{steptool} perform step-level credit assignment using DPO~\citep{dpo} and PPO~\citep{ppo}, respectively. TTPA~\citep{TTPA} further refines the granularity of preference alignment to the token level. Some works, such as~\citep{parallel_tool}, extend sequential tool learning to parallel settings, enabling agent tools to invoke multiple tools simultaneously. Moreover, with the recent advances in reinforcement learning using verifiable rewards~\citep{deepseek-r1, tuluv3}, ToolRL~\citep{toolrl} introduces tool-oriented reward design and achieves superior performance. Despite these advances, these flat tool-calling methods face a critical limitation: when handling long-horizon tasks that require numerous tools, both the intermediate task-level decision-making process and tool invocations continuously expand the context window, leaving the planner overloaded and suffering from planning burden and error accumulation.

\paragraph{Tool-use Ability Evaluation} 
Early studies such as APIBench~\citep{apibench} and ToolBench v1~\citep{xu2023tool} primarily focus on the tool learning preparation stage, evaluating the correctness of tool usage and parameter selection. Subsequent benchmarks, including RestBench~\citep{restgpt}, T-Eval~\citep{chen2023t}, and API-Bank~\citep{api-bank} decouple the assessment of LLMs’ tool-learning capabilities into more fine-grained tasks, such as instruction following, task planning, format correctness, tool selection, and others. Additionally, ToolBench v2~\citep{toolllm} introduces a more comprehensive multi-tool benchmark, offering a thorough evaluation suite for assessing the real-world tool-use abilities of LLMs. Recently, the Berkeley Function-Calling Leaderboard (BFCL)~\citep{bfcl} has been widely adopted to evaluate tool-augmented agents. BFCL covers a broad range of challenges, including single-step reasoning, multi-step tool-use, real-time execution, rejection of irrelevant tools, simultaneous selection of multiple tools, and coordinated multi-tool execution. Although these benchmarks provide a comprehensive evaluation of different capabilities in tool-use, they mainly focus on short-horizon problems with limited complexity.

%% file: Sections/03-Method.tex
\section{Methodology} \label{method}
\subsection{Overview}
\begin{figure*}[!t]
        \centering
	\includegraphics[width=1\linewidth]{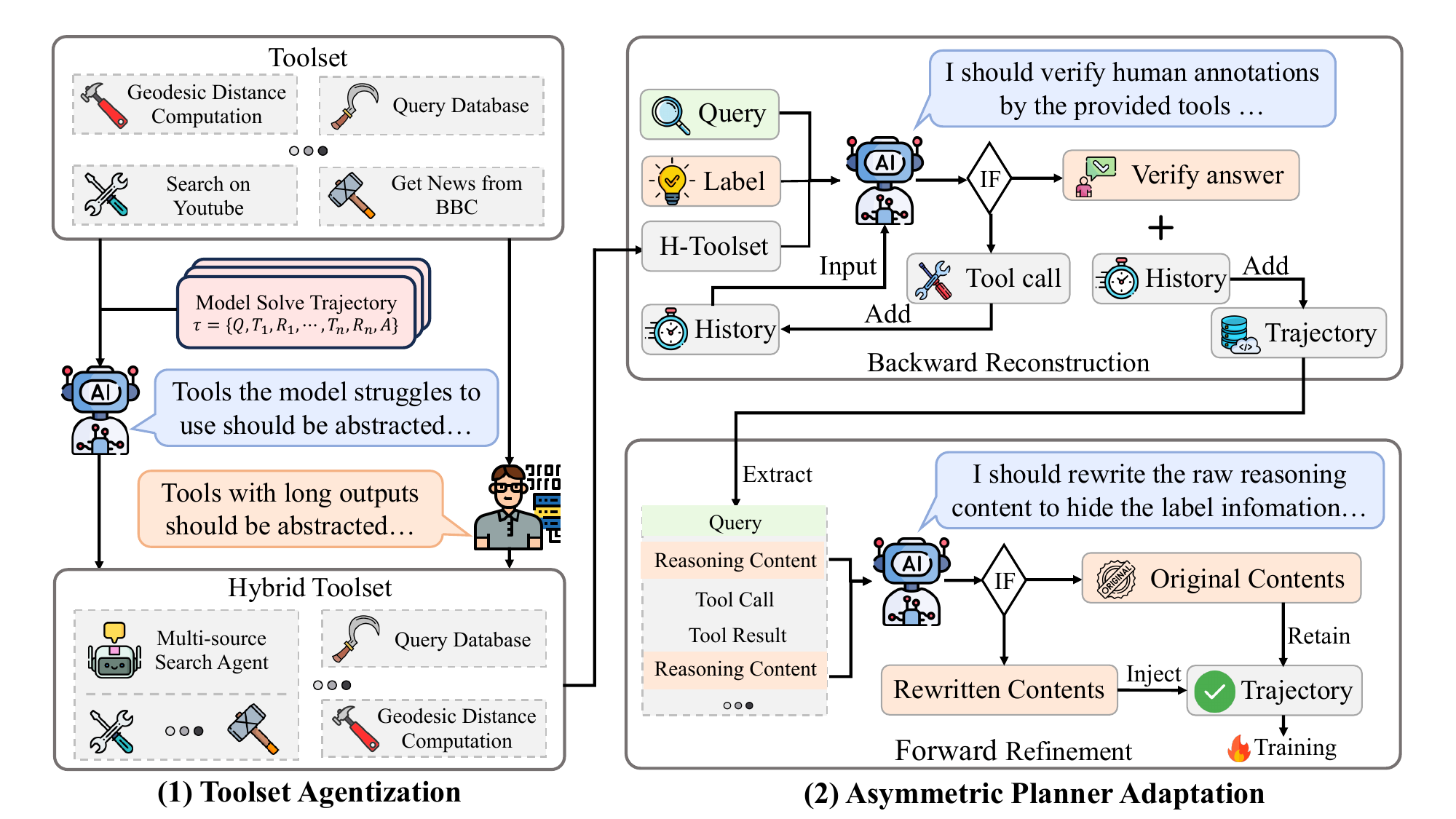}
  \caption{Overview of the \model framework. The methodology comprises two core components: \textbf{Toolset Agentization}, which abstracts fine-grained tools into a hybrid toolset (combining basic and agent tools) to reduce planning complexity; and \textbf{Asymmetric Planner Adaptation}, which aligns the planner via a trajectory construction pipeline (backward reconstruction and forward refinement) followed by hybrid policy optimization, while keeping the agent tools frozen to ensure stable coordination.}
 \label{fig:method}
\end{figure*}

We present a two-stage framework, \fullmodel (\model), to improve scalability and coordination in tool-augmented planning, as illustrated in Figure~\ref{fig:method}. 
Firstly, we propose \textbf{toolset agentization}, which abstracts groups of fine-grained tools into higher-level agent tools, allowing the planner to delegate complex subtasks.
Under this hybrid framework, the planner can not directly observe the outputs of tools, which leads to coordination consistency issues for the planner.
To address this issue, we further propose \textbf{asymmetric planner adaptation}, a trajectory-based training paradigm that enables the planner to robustly align with these agent tools. 

\subsection{Toolset Agentization}

Existing tool-augmented planners often rely on long chains of fine-grained function calls, where each tool invocation must be explicitly scheduled and interpreted by the planner. 
As the number of tools grows, such flat compositions substantially increase the planning horizon and cognitive burden, making high-level reasoning brittle and error-prone.
To address this issue, we propose toolset agentization, which abstracts groups of related tools into higher-level agent tools that encapsulate low-level executions behind a simplified interface. 
By delegating complex subtasks to these agents, the planner can operate over a shorter and more structured action space, focusing on high-level decision making rather than detailed tool coordination.

Importantly, not all tools benefit equally from agentization. 
Simple and deterministic tools already provide stable and lightweight function calls, and converting them into agents may unnecessarily introduce latency and behavioral variance. 
We therefore adopt a selective agentization strategy and maintain a hybrid toolset consisting of both agent tools and basic tools.
Specifically, we consider two complementary strategies for identifying tools to agentize. 
Let $\mathcal{T} = {t_1, \dots, t_N}$ denote the set of available tools.
Firstly, we leverage an LLM to analyze the available toolset and automatically select tools that are frequently invoked, strongly task-relevant, or prone to planner errors, as these tools typically require complex coordination and thus benefit most from abstraction. Concretely, we assign each tool a utility score $s(t)$ and select a subset: 
\begin{equation}
\mathcal{T}_{\text{sel}} = \{ t \in \mathcal{T} \mid s(t) > \tau \}
\end{equation}
where $s(t)$ denotes the LLM-estimated utility score and $\tau$ is a threshold. 
All tools not selected for agentization constitute the basic tool set:
$
\mathcal{T}_{\text{basic}} = \mathcal{T} \setminus \mathcal{T}_{\text{sel}}.
$
Secondly, we manually group some specific tools into clusters:
\begin{equation}
\{\mathcal{G}_1, \dots, \mathcal{G}_K\}, \quad
\bigcup_{k=1}^{K} \mathcal{G}_k \subseteq \mathcal{T}
\end{equation}
where each group $\mathcal{G}_k$ contains tools that have similar functionalities. 
Although these similar tools may access different information sources and provide complementary evidence, their outputs are often largely overlapping and differ mainly in emphasis. 
Invoking them independently forces the planner to process multiple lengthy and partially redundant responses, which increases context length and diffuses the planner’s attention across repetitive details, ultimately degrading reasoning performance. 
For each group $\mathcal{G}_k$, we construct an agent tool:
\begin{equation}
    a_k(x) = \mathrm{Agg}\big(\{ t(x) \mid t \in \mathcal{G}_k \}\big)
\end{equation}
where $\mathrm{Agg}(\cdot)$ denotes an internal aggregation and summarization operator.
By clustering such related capabilities, the agent can internally aggregate and summarize multi-source information, reducing redundancy and presenting a more concise and structured interface. 
Overall, toolset agentization yields a hybrid toolset that combines both basic tools and agent tools. 
Let the hybrid toolset be
\begin{equation}
    \mathcal{H} = \mathcal{T}_{\text{basic}} \cup \mathcal{A},
\end{equation}
where $\mathcal{A}=\{a_1,\dots,a_K\}$ denotes the set of agent tools constructed from selected clusters.
Given an input query $q$, the planner parameterized by $\pi_\theta$ selects an action $c = \pi_\theta(q, \mathcal{H})$
where $c$ corresponds to either a tool call or an agent call. 
The tool call or agent call's returned result is computed as
\begin{equation}
r =
\begin{cases}
t(q), & \text{if } c = t \in \mathcal{T}_{\text{basic}}, \\
\mathrm{Agg}(\{t_i(q)\}_{t_i \in \mathcal{G}_k}), & \text{if } c = a_k \in \mathcal{A}.
\end{cases}
\end{equation}
where agent tools internally orchestrate multiple tool invocations and aggregate their outputs. This unified formulation enables the planner to reason over a shorter action horizon while delegating low-level coordination to agents.
When the planner integrates results from multiple tool or agent calls to produce the final answer:
\begin{equation}
    y = \mathrm{LLM}(q, r_{i:m})
\end{equation}
where $r_{1:m}$ denotes the set of agent and basic tools' results collected during planning.

With toolset agentization, instead of explicitly invoking and coordinating individual tools, the planner delegates an entire subtask to a single agent tool. 
The agent autonomously determines the execution order of internal tools, manages their interactions, and consolidates intermediate outputs into a concise response. 
From the planner’s perspective, each agent tool behaves as a single callable basic tool with a well-defined interface, effectively hiding low-level complexity while preserving modularity. 

\subsection{Asymmetric Planner Adaptation}
To train the planner within \model, we adopt an asymmetric supervision strategy that explicitly addresses the coordination gap between the planner and the agent tools. 
Rather than relying solely on raw model generations, we construct high-quality training trajectories through a two-stage pipeline consisting of backward trajectory construction followed by forward trajectory refinement. 
Specifically, we first reconstruct tool-use trajectories from labeled outcomes to efficiently obtain feasible reasoning paths with high success rates. 
We then further refine these trajectories using a stronger teacher model to improve their logical coherence, semantic consistency, and tool-calling accuracy. 
The resulting trajectories provide reliable supervision signals for both imitation learning and subsequent policy optimization, enabling the planner to better align with the behaviors of agent tools and achieve more stable coordination.

\subsubsection{Backward Reconstruction}

Existing datasets for tool-augmented reasoning typically provide only input queries and final labels, while explicit tool-use trajectories are rarely available. 
Let $\mathcal{D} = \{(q_i, y_i)\}_{i=1}^M$ denote the dataset of queries and labels, where each trajectory 
$\tau = (c_1, r_1, \dots, c_T, r_T)$ consists of a sequence of tool or agent calls $c_t$ and their corresponding results $r_t$. 
As a result, constructing high-quality trajectories $\tau$ for supervision becomes a critical challenge for training tool-aware planners.
A common practice is to directly prompt an LLM to solve each query and sample trajectories from its policy
\begin{equation}
    \tau \sim \pi_\theta(\cdot \mid q),
\end{equation}
and then retain only those whose final prediction $\hat{y}(\tau)$ matches the ground-truth label $y$. 
However, this forward generation paradigm is often highly inefficient. 
When the model's task-specific capability is limited—particularly in domains with scarce pretraining signals such as POI verification, it frequently fails to produce correct solutions, resulting in a very low yield of usable trajectories. 
Consequently, a large number of generations must be sampled and repeatedly filtered, leading to substantial computational overhead and unstable data quality.
To address this issue, we adopt a \emph{backward reconstruction} strategy. 
Instead of generating solutions from scratch, we condition the model on both the query and the ground-truth label and construct trajectories from the conditional policy
\begin{equation}
    \tau \sim \pi_\theta(\cdot \mid q, y),
\end{equation}
prompting the model to verify or justify the given answer. 
Under this setting, the model performs reasoning with access to the correct outcome, allowing it to infer which tools should be invoked and in what order to support the label. 
As the model validates the label, it simultaneously produces a coherent sequence of tool calls that constitutes a valid trajectory.
This backward formulation substantially improves trajectory quality and yield. 
Intuitively, deriving the reasoning process given the correct outcome is easier than discovering both the process and the answer simultaneously, enabling more stable and efficient supervision for subsequent training. 
From a search perspective, conditioning on $y$ effectively reduces the solution space from $p(\tau \mid q)$ to $p(\tau \mid q, y)$, thereby alleviating exploration difficulty.
This design decouples trajectory feasibility from trajectory quality, allowing each stage to specialize in one objective.

\subsubsection{Forward Refinement}
Although backward reconstruction efficiently produces feasible and label-consistent tool trajectories, the resulting reasoning process is inherently conditioned on the ground-truth outcome. 
Consequently, these trajectories often reflect a \emph{verification-style} reasoning pattern—i.e., justifying a known answer—rather than the \emph{forward problem-solving} behavior exhibited during real inference. 
Such discrepancies may introduce distribution mismatch between training and deployment, potentially limiting the planner’s generalization and coordination with tools.
To mitigate this gap, we introduce a forward refinement stage that converts reconstructed trajectories into more natural, forward-executable reasoning processes. 
Specifically, given a reconstructed trajectory $\tau$, we employ a stronger teacher model to refine the associated reasoning process while preserving the original tool execution structure. 
Instead of regenerating new tool calls, we keep the sequence of calls and results fixed, and prompt the teacher to rewrite the intermediate reasoning steps and the final answer in a forward problem-solving manner conditioned only on the original query $q$. 
This design is motivated by the observation that backward reconstruction already yields correct and executable tool trajectories, since verifying the label implicitly solves the task. 
Therefore, modifying the execution plan is unnecessary and may introduce additional noise.
Formally, we perform a structure-preserving refinement
\begin{equation}
    \tilde{\tau} = \mathrm{Refine}_{\text{teacher}}(\tau \mid q),
\end{equation}
where the tool calls remain unchanged, and only the reasoning chain and answer synthesis are improved. 
The teacher enhances logical coherence, removes label-dependent or verification-style shortcuts, and reorganizes the explanations to reflect a causal, forward reasoning process.
This refinement step aligns training trajectories with the planner’s test-time behavior, improving semantic clarity and executability without sacrificing correctness. 
Combined with backward reconstruction, our two-stage pipeline decouples trajectory feasibility from trajectory quality: the former guarantees valid tool usage, while the latter improves reasoning naturalness and policy alignment. 
The resulting refined trajectories $\tilde{\tau}$ are aggregated to form the training dataset
\begin{equation}
\mathcal{D}_{\text{train}} = \{(q_i, \tilde{\tau}_i)\}_{i=1}^M,
\end{equation}
which is subsequently used for supervised fine-tuning and policy optimization of the planner.

\subsubsection{Hybrid Policy Optimization}\label{sec:optimization}

To align the planner with the refined expert demonstrations, we employ a supervised optimization framework based on \textit{Behavioral Cloning}. This design provides a stable initialization by learning from high-quality trajectories constructed via the asymmetric adaptation pipeline, avoiding the instability introduced by reinforcement learning in high-dimensional action spaces.
We initialize the planner via SFT on the refined trajectory dataset $\mathcal{D}_{\text{train}}$. This phase constrains the policy manifold to regions of high feasibility, encouraging the model to imitate effective coordination patterns between basic tools and agent tools. 
Then the planner learns to replicate the coordination strategies demonstrated in expert trajectories, ensuring that the initial policy lies in a region that supports valid and efficient tool execution. 
Starting from curated trajectories significantly reduces exploration difficulty and improves training stability.
A critical feature of our framework is the \textit{asymmetric} nature of the optimization. During training, the parameters of the underlying agent tools and basic tools remain frozen; only the planner parameters $\theta$ are updated. Formally, this treats the toolset $\mathcal{H}$ as a static environment $\mathcal{E}$, rendering the transition dynamics $P(r_t \mid c_t, \mathcal{E})$ stationary from the planner's perspective. This constraint stabilizes training by preventing non-stationarity induced by co-adapting tools, allowing the planner to focus on learning robust coordination strategies over a fixed set of capabilities.

%% file: Sections/04-Experiments.tex
\section{Experiments}

\subsection{Datasets}

\paragraph{\dataname}
We argue that it is crucial to assess the aforementioned planning and tool-use abilities in realistic settings. 
Thus, we conduct experiments on a real-world verification dataset with 700+ tasks and 15+ tools.
In this dataset, LLMs must invoke multiple tools to retrieve information from diverse sources, collect supporting evidence, and verify the correctness of given claims. 
The curation process of this dataset proceeds as follows: 
(1) We begin with real-world verification tasks collected from our industrial scenarios, each accompanied by human-verified ground-truth outcomes. 
These instances are split into training and test sets in a 9:1 ratio. 
(2) We analyze and distill the manual workflows performed by human operators into a set of well-defined, reusable tools. 
Each tool represents an atomic, executable action that can be invoked during reasoning. 
(3) Leveraging the original verification inputs and their corresponding human-validated results, we adopt Claude-Sonnet-4.5 to construct prior-informed trajectories. 
(4) We prompt Gemini-3-Flash-Preview to reconstruct these trajectories into natural, self-consistent reasoning processes that arrive at the correct solution without leveraging the known answer, effectively simulating how an agent would solve the query from scratch. 
These de-prioritized and high-quality trajectories form the basis of our SFT data.
The dataset encompasses multiple categories of verification challenges, including those involving conflicting or redundant information, etc. 
For training and validate on this dataset, we further sample 5000+ trajectories for SFT, where around 5000 are used for training and the remaining 200+ for validation.
Subsequently, we fine-tune Qwen3-8B\citep{qwen3} as the planner on the constructed dataset to optimize its performance. More implementation details can be found in Appendix~\ref{app:details}.

\paragraph{Extended BFCL}
Existing work has primarily utilized the BFCL to evaluate the tool-use capabilities of LLMs. 
Despite its advances in containing a wide range of challenges, the longest reasoning trace for any question in BFCL comprises at most only five steps, which is insufficient to evaluate the long-horizon planning and tool-use abilities of LLMs. 
To address this limitation, we synthesize a set of complex queries based on BFCL’s multi-turn configurations, where each query requires an average of 9.4 rounds of tool invocations to resolve. 
We synthesize the Extended BFCL with 200 tasks through the following pipeline: 
(1) Compute the Cartesian product of simple simulated environment configurations, resulting in complex environments. 
(2) Use a powerful LLM to generate queries for each of the above configurations, retaining only those combinations of environments and queries that are reasonable and natural. 
(3) Apply rejected sampling to the aforementioned queries, employ the LLM-as-a-judge along with pass@k to filter challenging queries, and obtain ground-truth answers from the sampling trajectories. 
This extended dataset includes both single-scenario and multi-scenario compositions, enabling a more reliable and accurate evaluation of LLMs’ capabilities across diverse settings.

\subsection{Main Results} 
\input{Tables/production-line}

\paragraph{\dataname}
As shown in Table~\ref{tab:prod-line}, native LLMs achieve only moderate performance (around 34.86\% accuracy), revealing limitations in handling long-horizon, tool-intensive tasks.
With \model, most models improve significantly: Claude-Sonnet-4.5 rises from 21.92\% to 35.06\%, GPT-4o gains 1.98\%, Qwen3-8B improves by 1.52\%, and ToolACE-2-8B by 8.37\%. 
DeepSeek-Chat shows comparable overall performance but benefits across subtasks. In contrast, xLAM-2-8B-FC-r slightly declines, likely due to its specialization in simulated tool-use, which limits real-world generalization. 
Additionally, results from APA-8B indicate that task-specific fine-tuning yields further gains beyond direct adaptation, highlighting its importance for effective tool-use. 
We attribute this improvement to improved coordination between the planner and agent tools.

\input{Tables/bfcl}
\paragraph{Extended BFCL}

In this section, we present the results on our extended BFCL dataset. As shown in Table~\ref{tab:bfcl_results}, although this extended tool-use dataset is simulated and the tools are relatively simple, most existing models still achieve limited performance. We argue that current LLMs suffer from issues such as cognitive overload and difficulty in mastering multiple tools, which prevents them from successfully completing long-horizon tasks. Despite these challenges, \model delivers a solid boost in performance, with improvements ranging from 0.5\% to 10.00\%. This advancement benefits from the hierarchical nature of \model, which separates the sequential and suffocating problem-solving process into high-level decision-making and low-level subtask execution.

\input{Tables/onlinetest}
\subsection{Scaling Abilities of \model}

To investigate \model’s scaling potential, we evaluate planners of different sizes (Qwen3-1.7B, 4B, 8B, 14B). As shown in Figure~\ref{fig:scaling_analysis}, larger planners achieve higher accuracy, showing clear scaling benefits.
We further study tool scaling using the Qwen3-8B series and DeepSeek-V3. With a fixed planner, performance improves as tool model size increases. Notably, a small planner with stronger tools can match larger planners, while \model consistently outperforms vanilla LLMs.
These results highlight the scalability and generalization of \model. Toolset agentization also shortens trajectories and improves token efficiency by reducing cognitive burden, enabling more effective planning.

\subsection{Efficiency Analysis}

In this section, we analyze the mechanism of \model, focusing on cognitive burden reduction via toolset agentization, measured by trajectory length and token usage. As shown in Figure~\ref{fig:traj_length} and Table~\ref{tab:token_analysis}, \model achieves shorter trajectories and higher token efficiency than variants without toolset agentization. By decomposing subtasks into specialized tools, \model reduces cognitive load and error accumulation, enabling the planner to focus on core decisions and avoid fragile reasoning paths. 
\subsection{Ablation Study}\label{sec:ablation_study}

In this section, we analyze the effectiveness of each key component in \model through the main results in Table~\ref{tab:prod-line}. Specifically, we compare \model with variants without toolset agentization (\model w/o TA) and without the dedicated training process (\model w/o APA). The comparisons in the main figure clearly demonstrate that both components contribute significantly to performance gains. 

\input{Tables/token_analysis}
\subsection{Online POI Information Verification Test}
In this section, we present a real-world industrial deployment of \model for POI information verification, processing tens of thousands tasks daily, where each task involves multi-step tool use. 
Given a candidate POI pair, the system first determines whether the two POIs are duplicates via external tools. 
If so, additional tools are invoked to verify their existence and validity. 
Based on multi-source signals, the system then resolves conflicts and selects an operation for the target POI, including keeping it unchanged, updating it, removing it (offline), or replacing it via a remove-then-add operation. 
We evaluate \model using industrial metrics, including automation rate (AR), daily task load rate (DTLR) per person, and number of staff required rate (NSRR).
Results in Table~\ref{tab:online_test} show that \model reduces human workload while improving automation and robustness in the production pipeline. 

%% file: Tables/production-line.tex
\begin{table}[htbp]
\centering
\caption{
Accuracy comparison on our proposed real-world information verification task. \textit{APA} indicates Qwen3-8B trained with Asymmetric Planner Adaptation. 
\textbf{Error}: incorrect POI attributes;
\textbf{Outdated}: obsolete or expired data;
\textbf{Missing}: missing POI information;
\textbf{Anomaly}: abnormal user behavior signals;
\textbf{Relocation}: POIs that have moved to new locations.
}
\begin{adjustbox}{width=\columnwidth,center}
\begin{tabular}{@{} p{4cm} >{\centering\arraybackslash}p{1.6cm} >{\centering\arraybackslash}p{1.6cm} >{\centering\arraybackslash}p{1.6cm} >{\centering\arraybackslash}p{1.6cm} >{\centering\arraybackslash}p{1.6cm} >{\centering\arraybackslash}p{1.6cm}@{}}
    \toprule
    \textbf{Models} & \textbf{Error} & \textbf{Outdated} & \textbf{Missing} & \textbf{Anomaly} & \textbf{Relocation} & \textbf{Avg.} \\ 
    \hline
    \multicolumn{7}{l}{\cellcolor{gray!20}\textit{General LLMs}}               \\
\textbf{Claude-Sonnet-4.5   }        & 13.07\%                                       & 33.11\%                                       & 39.00\%                                       & 13.59\%                                      & 14.00\%                                      & 21.92\%                           \\ 
\textbf{Claude-Sonnet-4.5 w. TA}  & 23.53\%                                       & 31.79\%                                       & 53.00\%                                       & 30.39\%                                      & 41.33\%                                      & \textbf{35.06\%}                           \\ 
\textbf{GPT-4o}             & 19.61\%                                       & 37.75\%                                       & 34.00\%                                       & 33.98\%                                      & 26.67\%                                      & 29.83\%                           \\ 
\textbf{GPT-4o w. TA}    & 24.18\%                                       & 39.74\%                                       & 31.00\%                                       & 32.04\%                                      & 32.00\%                                      & \textbf{31.81\%}                           \\ 
\textbf{Deepseek-chat }              & 28.76\%                                       & 43.05\%                                       & 37.00\%                                       & 22.33\%                                      & 40.00\%                                      & 34.86\%                           \\ 
\textbf{Deepseek-chat w. TA}      & 32.03\%                                       & 39.07\%                                       & 43.00\%                                       & 26.21\%                                      & 34.00\%                                      & 34.86\%                           \\ 

\textbf{Qwen3-8B}           & 25.49\%                                       & 39.07\%                                       & 34.00\%                                       & 22.33\%                                      & 38.00\%                                      & 32.27\%                           \\ 
\textbf{Qwen3-8B w. TA}  & 30.07\%                                       & 35.76\%                                       & 31.00\%                                       & 25.24\%                                      & 43.33\%                                      & \textbf{33.79\%}                           \\ 
    \hline
    \multicolumn{7}{l}{\cellcolor{gray!20}\textit{Tool-augmented LLMs}}               \\
\textbf{ToolACE-2-8B}          & 23.53\%                                       & 16.56\%                                       & 23.00\%                                       & 33.01\%                                      & 28.67\%                                      & 24.51\%                           \\ 
\textbf{ToolACE-2-8B w. TA} & 24.18\%                                       & 32.45\%                                       & 37.00\%                                       & 44.66\%                                      & 31.33\%                                      & \textbf{32.88\%}                          \\ 
\textbf{xLAM-2-8B-FC}               & 13.73\%                                       & 19.87\%                                       & 15.00\%                                       & 31.07\%                                      & 25.33\%                                      & \textbf{20.70\%}                           \\ 
\textbf{xLAM-2-8B-FC w. TA}      & 9.80\%                                        & 19.21\%                                       & 10.00\%                                       & 33.98\%                                      & 30.67\%                                      & 20.55\%      
\\
\textbf{APA-8B}          & 49.02\% & 52.98\% & 51.00\% & 48.54\% & 51.33\% & 50.68\% \\
\textbf{APA-8B w. TA} & 53.59\% & 51.66\% & 57.00\% & 55.34\% & 53.33\% & 53.88\% \\

    \bottomrule
\end{tabular}
\end{adjustbox}

\label{tab:prod-line}
\end{table}

%% file: Tables/bfcl.tex

\begin{table*}[htbp]
\centering
\caption{
Accuracy comparison on Extended BFCL. Whether strong closed-source models or specialized tool-use models, TA significantly boosts their performance. 
}
\begin{adjustbox}{width=0.98\columnwidth,center}
\begin{tabular}{@{} p{4cm} >{\centering\arraybackslash}p{0.9cm} >{\centering\arraybackslash}p{0.9cm} >{\centering\arraybackslash}p{0.9cm} >{\centering\arraybackslash}p{0.9cm} >{\centering\arraybackslash}p{0.9cm} >{\centering\arraybackslash}p{0.9cm} >{\centering\arraybackslash}p{0.9cm} >{\centering\arraybackslash}p{0.9cm} >{\centering\arraybackslash}p{0.9cm}@{}}
    \toprule
    {\multirow{2}{*}{\textbf{Models}}}
    & \multicolumn{4}{c}{\textbf{Single}} 
    & \multicolumn{4}{c}{\textbf{Multiple}} 
    & \multirow{2}{*}{\textbf{Total}} \\
    \cmidrule(lr){2-5} \cmidrule(lr){6-9} 
    & \textbf{File} & \textbf{Vehicle} & \textbf{Trade} & \textbf{Travel} & \textbf{File} & \textbf{Vehicle} & \textbf{Trade} & \textbf{Travel} & \textbf{} \\
    \hline
    
    \multicolumn{10}{l}{\cellcolor{gray!20}\textit{General LLMs}}        \\
    
    \textbf{Claude-Sonnet-4.5} &53.85\%	&52.63\%	&85.00\%	&84.62\%	&78.38\%	&67.74\%	&70.00\%	&67.57\%	&70.50\% \\
    \textbf{Claude-Sonnet-4.5 w. TA} &61.54\%	&52.63\%	&85.00\%	&84.62\%	&75.68\%	&70.97\%	&73.33\%	&64.86\%	&\textbf{71.00}\% \\  
    \textbf{GPT-4o} &38.46\%	&47.37\%	&90.00\%	&92.31\%	&48.65\%	&64.52\%	&66.67\%	&40.54\%	&58.50\% \\    
    \textbf{GPT-4o w. TA} &38.46\%	&47.37\%	&95.00\%	&100.0\%	&51.35\%	&64.52\%	&66.67\%	&45.95\%	&\textbf{61.00}\% \\   
    \textbf{DeepSeek-Chat} &53.85\%	&15.79\%	&55.00\%	&61.54\%	&40.54\%	&16.13\%	&46.67\%	&40.54\%	&39.00\% \\    
    \textbf{DeepSeek-Chat w. TA} &46.15\%	&5.26\%	&70.00\%	&30.77\%	&56.76\%	&12.90\%	&56.67\%	&32.43\%	&\textbf{39.50}\% \\   
    \textbf{Qwen3-8B} &23.08\%	&10.53\%	&65.00\%	&23.08\%	&32.43\%	&38.71\%	&56.67\%	&27.03\%	&36.00\% \\  
    \textbf{Qwen3-8B w. TA} &23.08\%	&21.05\%	&65.00\%	&30.77\%	&37.84\%	&32.26\%	&60.00\%	&32.43\%	&\textbf{39.00}\% \\ 
    
    \hline
        \multicolumn{10}{l}{\cellcolor{gray!20}\textit{Tool-augmented LLMs}}               \\
    \textbf{ToolACE-2-8B} &15.38\%	&21.05\%	&45.00\%	&46.15\%	&16.22\%	&19.35\%	&43.33\%	&16.22\%	&26.00\% \\    
    \textbf{ToolACE-2-8B w. TA} &30.77\%	&21.05\%	&65.00\%	&46.15\%	&29.73\%	&29.03\%	&53.33\%	&24.32\%	&\textbf{36.00}\% \\  
    \textbf{xLAM-2-8B-FC} &38.46\%	&31.58\%	&50.00\%	&53.85\%	&59.46\%	&22.58\%	&66.67\%	&45.95\%	&47.00\% \\    
    \textbf{xLAM-2-8B-FC w. TA} &46.15\%	&26.32\%	&50.00\%	&69.23\%	&59.46\%	&22.58\%	&66.67\%	&48.65\%	&\textbf{48.50}\% \\  
    \bottomrule
\end{tabular}
\end{adjustbox}

\label{tab:bfcl_results}
\end{table*}


%% file: Tables/onlinetest.tex
\begin{figure}[h]
    \centering
    \begin{minipage}{0.47\linewidth}
        \centering
        \includegraphics[width=\linewidth]{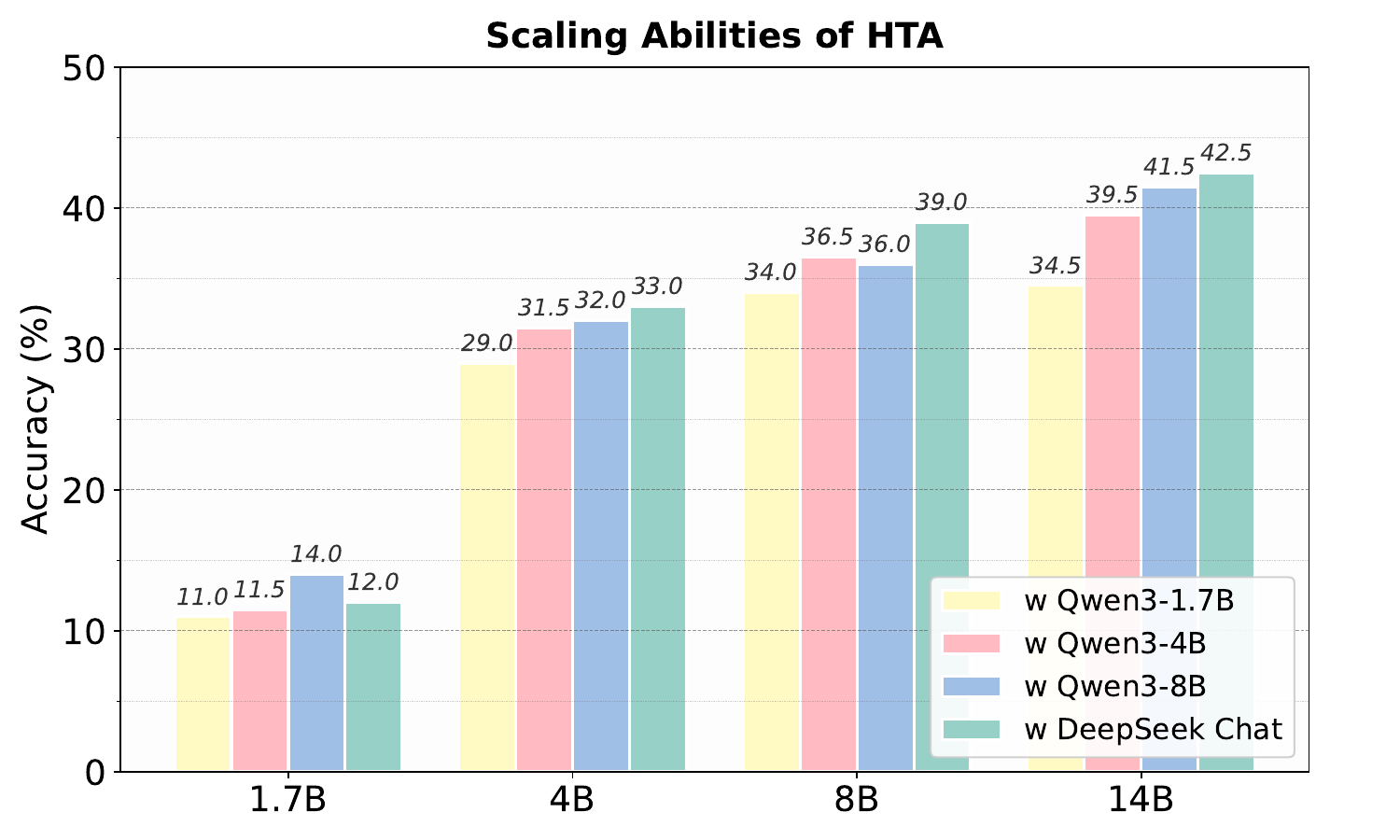} 
        \caption{Scaling experiments of \model with varying planner sizes and tool agent sizes.} 
        \label{fig:scaling_analysis} 
    \end{minipage}%
    \hfill 
    \begin{minipage}{0.5\linewidth}
        \centering
        \includegraphics[width=\linewidth]{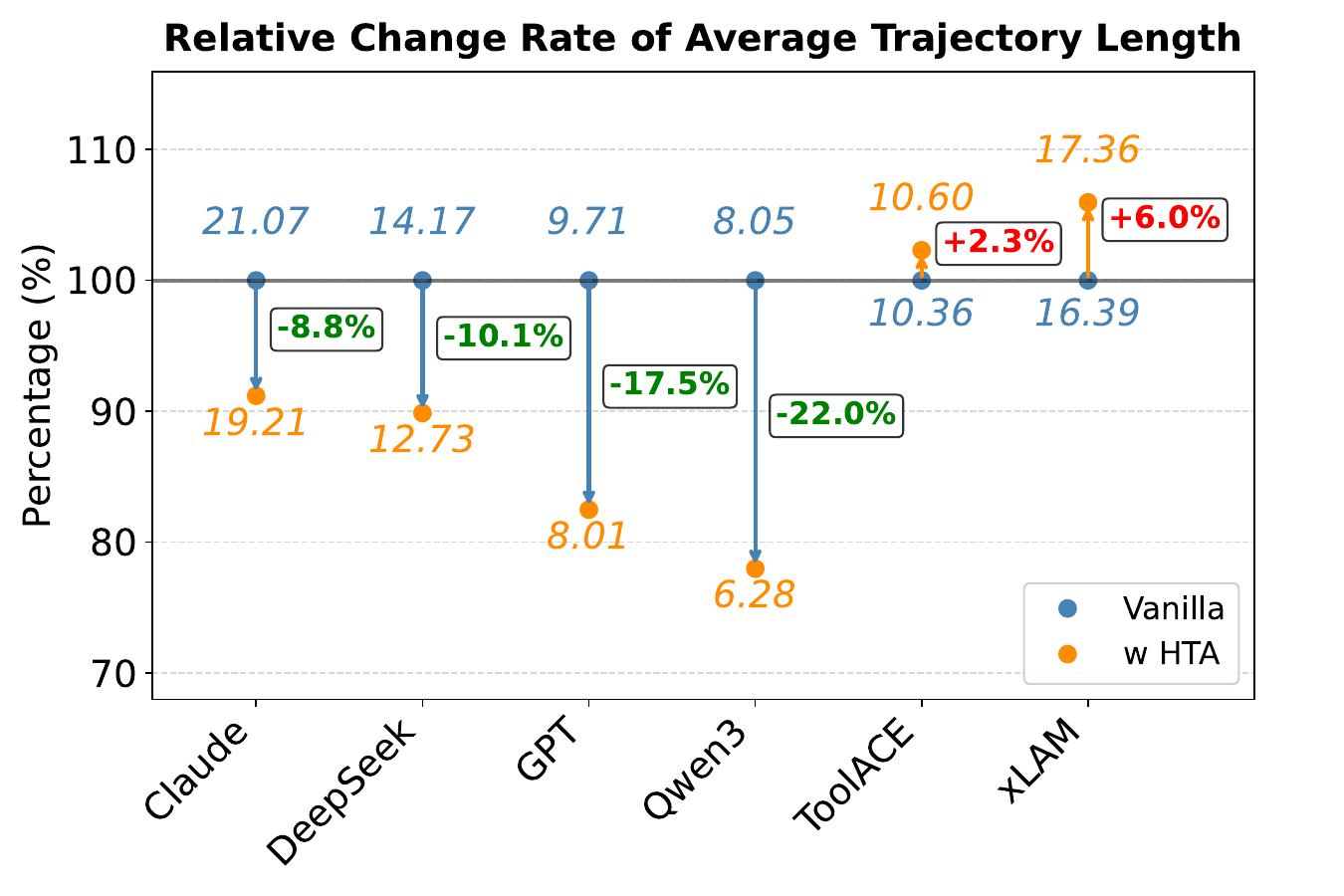} 
        \caption{Trajectory length comparison across different models.} 
        \label{fig:traj_length} 
    \end{minipage}%
\end{figure}


%% file: Tables/token_analysis.tex

\begin{table}[!htbp]
    \centering
    \begin{minipage}{0.43\linewidth}
        \centering
        \fontsize{8}{11}\selectfont
                \caption{Comparison of manual, rule-based, and HTA approaches on industrial POI verification tasks.}
        \begin{tabular}{@{} p{1.4cm} >{\centering\arraybackslash}p{1cm} >{\centering\arraybackslash}p{1cm} >{\centering\arraybackslash}p{1cm} @{}}
        \toprule 
        \textbf{Models} 
        & \textbf{AR} 
        & \textbf{DTLR}
        & \textbf{NSRR} \\  
        \midrule
        Manual      & 0.00\% & 100\% & 100\%  \\
        Rule-Based     & 30.0\% & 69.99\% & 81.25\%  \\
        \model   & 84.5\% & 15.49\% & 18.75\%  \\
        \bottomrule 
        \end{tabular} 

        \label{tab:online_test}
    \end{minipage}
    \hfill 
    \begin{minipage}{0.53\linewidth}
        \centering
        \fontsize{8}{11}\selectfont
                \caption{Token usage analysis with and without the proposed method.}
        \begin{tabular}{@{} p{2.2cm} >{\centering\arraybackslash}p{1.4cm} >{\centering\arraybackslash}p{1.2cm} >{\centering\arraybackslash}p{1.2cm} @{}}
        \toprule 
        \textbf{Models} 
        & \textbf{w/o. \model} 
        & \textbf{w. \model}
        & \textbf{Eff. $\Delta$\%} \\  
        \midrule 
        Claude-Sonnet-4.5 & 18357.2 & 9073.0 & 50.6\% \\
        DeepSeek-Chat & 10483.0 & 6938.0 & 33.8\% \\
        GPT-4o & 7476.3 & 4106.4 & 45.1\% \\
        Qwen3-8B & 7087.0 & 4215.0 & 40.5\% \\
        ToolACE & 5989.2 & 5070.2 & 15.3\% \\
        xLAM & 9957.0 & 7454.2 & 25.1\% \\
        \bottomrule 
        \end{tabular} 

        \label{tab:token_analysis}
    \end{minipage}
\end{table}

%% file: Sections/05-Conclusion.tex
\section{Conclusion}
We propose \fullmodel (\model), a hierarchical framework that improves the scalability and reliability of tool-augmented LLM planning by abstracting tools into agents and aligning them with the planner via asymmetric adaptation. Experiments on both benchmark and real-world datasets demonstrate that \model achieves higher task success rates with shorter trajectories and lower context overhead, while practical deployment further confirms its effectiveness in reducing manual effort and operational costs, highlighting its potential for complex, long-horizon real-world applications.

%% file: Sections/06-Appendix.tex
\appendix
\newpage
\section{Implementation Details} \label{app:details}
We conduct experiments on a range of representative LLMs: (1) Closed-source LLMs: DeepSeek-Chat~\citep{deepseek_v3}, GPT-4o~\citep{gpt4} and Claude-Sonnet-4.5, which demonstrate strong capabilities in tool usage and agentic tasks. (2) Tool-use expert models: ToolACE-2-8B~\citep{liu2024toolacewinningpointsllm} and xLAM-2-8B-FC~\citep{xlam}, both of which are open-weight LLMs specifically fine-tuned for function calling and multi-tool coordination. (3) Open-sourced Qwen3-Series~\citep{qwen3}, which exhibit superior performance across a wide range of natural language tasks.

For \textit{hybrid toolset agentization}, we employ GPT-5.2 to manage low-level tool orchestration and response aggregation, constructing the hierarchical tool ecosystem. 
For training and validate on this dataset, we further sample 5342 trajectories for SFT, where 5075 are used for training and the remaining 269 for validation.
Subsequently, we fine-tune Qwen3-8B\citep{qwen3} as the planner on the constructed dataset to optimize its performance. 
The details of training hyper-parameters can be found in Table~\ref{tab:train}. 
All training experiments are conducted on a single machine with 8×A100 GPUs (80GB memory each). Inference is performed on a separate machine equipped with 1×A100 GPU (80GB memory). The detailed pipeline for hybrid toolset construction and trajectory data generation is presented in Algorithm~\ref{alg:htaa_unified}.

\begin{algorithm}[htbp]
\caption{HTAA Unified Data \& Trajectory Construction}
\label{alg:htaa_unified}
\begin{algorithmic}[1]
\Require Tool set $\mathcal{T}=\{t_1,\dots,t_N\}$, Dataset $\mathcal{D}=\{(q_i, y_i)\}_{i=1}^M$, Planner $\pi_\theta$, Teacher model $\mathcal{T}_e$, Threshold $\tau$
\Ensure Hybrid toolset $\mathcal{H}$, Training dataset $\mathcal{D}_{train}$

\Comment{Stage 1: Toolset Agentization}
\For{each tool $t \in \mathcal{T}$}
    \State Compute utility score $s(t)$ using LLM
\EndFor
\State $\mathcal{T}_{sel} \gets \{t \in \mathcal{T} \mid s(t) > \tau\}$
\State $\mathcal{T}_{basic} \gets \mathcal{T} \setminus \mathcal{T}_{sel}$

\State Construct tool groups $\{G_1, \dots, G_K\}$ from $\mathcal{T}_{sel}$
\For{each group $G_k$}
    \State Define agent $a_k(x) \gets \mathrm{Agg}(\{t(x)\mid t \in G_k\})$
\EndFor
\State $\mathcal{A} \gets \{a_1,\dots,a_K\}$
\State $\mathcal{H} \gets \mathcal{T}_{basic} \cup \mathcal{A}$

\Comment{Stage 2: Trajectory Construction}
\State Initialize $\mathcal{D}_{train} \gets \emptyset$

\For{each $(q, y) \in \mathcal{D}$}
    \Comment{Backward Reconstruction}
    \State Sample trajectory $\tau \sim \pi_\theta(\cdot \mid q, y, \mathcal{H})$
    \State $\tau = (c_1, r_1, \dots, c_T, r_T)$
    \For{$t = 1$ to $T$}
        \If{$c_t \in \mathcal{T}_{basic}$}
            \State $r_t \gets c_t(q)$
        \Else
            \State $r_t \gets \mathrm{Agg}(\{t_i(q)\mid t_i \in G_k\})$
        \EndIf
    \EndFor

    \Comment{Forward Refinement}
    \State $\tilde{\tau} \gets \mathcal{T}_e(\tau \mid q)$
    \State Preserve $\{c_t, r_t\}_{t=1}^T$
    \State Rewrite reasoning steps into forward style

    \Comment{Collect Data}
    \State $\mathcal{D}_{train} \gets \mathcal{D}_{train} \cup \{(q, \tilde{\tau})\}$
\EndFor

\State \Return $\mathcal{H}, \mathcal{D}_{train}$
\end{algorithmic}
\end{algorithm}
\input{Tables/training-details}

\section{Limitations}\label{app:limitation}
While HTAA shows strong empirical performance, several limitations remain. The effectiveness of toolset agentization depends on the quality of tool grouping, which may vary across domains. In addition, abstracting tools into higher-level agents can reduce access to fine-grained intermediate information in certain scenarios. Finally, our evaluation includes a proprietary dataset, which may limit direct reproducibility, though we provide detailed descriptions to facilitate understanding.

\section{Broader Impact}\label{app:broder}
This work aims to improve the scalability and efficiency of tool-augmented large language models, with potential positive impacts on real-world applications that require complex decision-making, such as information verification, workflow automation, and intelligent assistants. By reducing reliance on manual operations, HTAA can improve productivity, lower operational costs, and enable more reliable large-scale deployment of tool-based systems.
More broadly, HTAA provides a practical framework for organizing large tool ecosystems, which may facilitate the development of more capable and structured agent systems in both research and industry. Its hierarchical design also offers a pathway toward improving robustness and efficiency in long-horizon reasoning tasks.
As with other LLM-based systems, there remain potential risks if the model produces incorrect decisions or is applied in high-stakes scenarios. In practice, these risks can be mitigated through standard safeguards such as human oversight, monitoring mechanisms, and appropriate deployment constraints.

%% file: Tables/training-details.tex
\begin{table}[h!]
    \caption{Hyper-parameters in experiments for training.}
    \centering
    \begin{adjustbox}{width=\columnwidth,center}
    \begin{tabular}{@{} *{7}{>{\centering\arraybackslash}m{2cm}} @{}}
    \toprule
         \textbf{Learning Rate}&  \textbf{Warm-up Ratio} &\textbf{LR Scheduler} &\textbf{Batch Size} & \textbf{valuation Ratio} & \textbf{bf16} &\textbf{Epochs}  \\ 
         \midrule
         $5\times10^{-5}$   & 0.05   & cosine    & 128 & 0.05 & true & 5 \\ 
    \bottomrule
     \end{tabular}
     \end{adjustbox}

    \label{tab:train}
\end{table}

%% file: checklist.tex
\section*{NeurIPS Paper Checklist}

\begin{enumerate}

\item {\bf Claims}
    \item[] Question: Do the main claims made in the abstract and introduction accurately reflect the paper's contributions and scope?
    \item[] Answer: \answerYes{} 
    \item[] Justification: The abstract and introduction clearly describe the proposed HTAA framework, including its key components (toolset agentization and asymmetric planner adaptation), main contributions, and targeted challenges. These claims are consistent with the experimental results presented in Section 4, which demonstrate improved task success rates, reduced trajectory length, and enhanced efficiency across benchmarks and real-world datasets. 

\item {\bf Limitations}
    \item[] Question: Does the paper discuss the limitations of the work performed by the authors?
    \item[] Answer: \answerYes{} 
    \item[] Justification: The limitations of the proposed method, including potential generalization constraints, abstraction trade-offs, and computational overhead, are discussed in Appendix ~\ref{app:limitation}.

\item {\bf Theory assumptions and proofs}
    \item[] Question: For each theoretical result, does the paper provide the full set of assumptions and a complete (and correct) proof?
    \item[] Answer: \answerNA{} 
    \item[] Justification: The paper does not present formal theoretical results such as theorems or proofs, but instead focuses on a methodological framework and empirical evaluation.

\item {\bf Experimental result reproducibility}
    \item[] Question: Does the paper fully disclose all the information needed to reproduce the main experimental results of the paper to the extent that it affects the main claims and/or conclusions of the paper (regardless of whether the code and data are provided or not)?
    \item[] Answer: \answerYes{} 
    \item[] Justification: The paper provides detailed implementation descriptions, including training procedures, trajectory construction, and pseudocode in the Appendix~\ref{app:details}, enabling reproducibility of the proposed method. Although the InfoVerify dataset is proprietary, its construction process and evaluation protocols are described to facilitate approximate replication. 

\item {\bf Open access to data and code}
    \item[] Question: Does the paper provide open access to the data and code, with sufficient instructions to faithfully reproduce the main experimental results, as described in supplemental material?
    \item[] Answer: \answerNo{} 
    \item[] Justification: The code and data are not publicly released. The InfoVerify dataset is proprietary and subject to access restrictions. However, the paper provides detailed descriptions of the methodology and experimental setup to facilitate understanding and partial reproducibility.

\item {\bf Experimental setting/details}
    \item[] Question: Does the paper specify all the training and test details (e.g., data splits, hyperparameters, how they were chosen, type of optimizer) necessary to understand the results?
    \item[] Answer: \answerYes{} 
    \item[] Justification: The paper specifies the experimental setup, including training procedures, model configurations, and evaluation protocols. Additional implementation details such as hyperparameters and data processing are provided in the Appendix~\ref{app:details}, enabling a clear understanding of the results. 

\item {\bf Experiment statistical significance}
    \item[] Question: Does the paper report error bars suitably and correctly defined or other appropriate information about the statistical significance of the experiments?
    \item[] Answer: \answerNo{} 
    \item[] Justification: The paper reports single-run results without error bars or formal statistical significance analysis. This is partly due to the high computational cost of large-scale LLM-based experiments, although consistent improvements are observed across multiple benchmarks.

\item {\bf Experiments compute resources}
    \item[] Question: For each experiment, does the paper provide sufficient information on the computer resources (type of compute workers, memory, time of execution) needed to reproduce the experiments?
    \item[] Answer: \answerYes{} 
    \item[] Justification: The paper provides information about the compute resources used, including hardware configuration, model types, and estimated training and inference costs, as described in the Appendix ~\ref{app:details}.
    
\item {\bf Code of ethics}
    \item[] Question: Does the research conducted in the paper conform, in every respect, with the NeurIPS Code of Ethics \url{https://neurips.cc/public/EthicsGuidelines}?
    \item[] Answer: \answerYes{} 
    \item[] Justification: The research complies with the NeurIPS Code of Ethics. It does not involve human subjects or sensitive personal data, and all experiments are conducted using publicly available benchmarks or proprietary datasets with appropriate usage considerations. 

\item {\bf Broader impacts}
    \item[] Question: Does the paper discuss both potential positive societal impacts and negative societal impacts of the work performed?
    \item[] Answer: \answerYes{} 
    \item[] Justification: The paper discusses both positive and negative societal impacts, including potential benefits in efficiency and risks related to misuse and incorrect automated decisions, as described in the Appendix~\ref{app:broder}.
    
\item {\bf Safeguards}
    \item[] Question: Does the paper describe safeguards that have been put in place for responsible release of data or models that have a high risk for misuse (e.g., pre-trained language models, image generators, or scraped datasets)?
    \item[] Answer: \answerNA{} 
    \item[] Justification: The paper does not release any datasets or models that pose a high risk of misuse. It focuses on a methodological framework and experimental evaluation, without distributing potentially sensitive resources. 

\item {\bf Licenses for existing assets}
    \item[] Question: Are the creators or original owners of assets (e.g., code, data, models), used in the paper, properly credited and are the license and terms of use explicitly mentioned and properly respected?
    \item[] Answer: \answerYes{} 
    \item[] Justification:  We have properly cited the original papers that provided the code packages and datasets used in our work. We have also ensured that the licenses and terms of use for these assets are fully respected.

\item {\bf New assets}
    \item[] Question: Are new assets introduced in the paper well documented and is the documentation provided alongside the assets?
    \item[] Answer: \answerNA{} 
    \item[] Justification: The paper does not release any new datasets, models, or code as public assets. While a proprietary dataset is used, it is not made publicly available.

\item {\bf Crowdsourcing and research with human subjects}
    \item[] Question: For crowdsourcing experiments and research with human subjects, does the paper include the full text of instructions given to participants and screenshots, if applicable, as well as details about compensation (if any)? 
    \item[] Answer: \answerNA{} 
    \item[] Justification: The paper does not involve crowdsourcing or research with human subjects. All experiments are conducted using automated evaluation on benchmark datasets and system outputs.

\item {\bf Institutional review board (IRB) approvals or equivalent for research with human subjects}
    \item[] Question: Does the paper describe potential risks incurred by study participants, whether such risks were disclosed to the subjects, and whether Institutional Review Board (IRB) approvals (or an equivalent approval/review based on the requirements of your country or institution) were obtained?
    \item[] Answer: \answerNA{} 
    \item[] Justification: The paper does not involve crowdsourcing or research with human subjects, and therefore does not require IRB approval or related ethical review.

\item {\bf Declaration of LLM usage}
    \item[] Question: Does the paper describe the usage of LLMs if it is an important, original, or non-standard component of the core methods in this research? Note that if the LLM is used only for writing, editing, or formatting purposes and does \emph{not} impact the core methodology, scientific rigor, or originality of the research, declaration is not required.
    \item[] Answer: \answerYes{} 
    \item[] Justification: The paper clearly describes the use of large language models as a central component of the proposed framework, including their roles in planning, tool invocation, and trajectory generation, along with the specific models used in experiments.

\end{enumerate}